\documentclass[a4paper, 10pt, conference]{ieeeconf}      

\IEEEoverridecommandlockouts                              

\usepackage{graphics} 
\usepackage{epsfig} 
\usepackage{mathptmx} 
\usepackage{times} 
\usepackage{booktabs}
\usepackage{multirow}
\usepackage{cite}
\usepackage{amsmath,amssymb,amsfonts}
\usepackage{algorithmic}
\usepackage{graphicx}
\usepackage{textcomp}
\usepackage{xcolor}
\usepackage{booktabs}
\usepackage{array}
\usepackage{multirow}
\usepackage{tabularx}
\usepackage{tcolorbox}
\usepackage{makecell}
\usepackage{algorithm}
\usepackage{algorithmic}
\title{\LARGE \bf
SOPD-SocialNav: Selective On-Policy Distillation for Vision-Language Social Navigation}

\author{Xinyu Zhang$^{1}$, Zishuo Wang$^{1}$, and Ling Xiao$^{1,a}$
\thanks{*This work was supported in part by JST Moonshot Goal 3 under Grant JPMJMS263E-12 and the NVIDIA Academic Grant Program.}
\thanks{$^a$Corresponding Author}%
\thanks{$^{1}$Graduate School of Information Science and Technology, Hokkaido University, Sapporo, Japan
 {\tt\small \{xinyu.zhang.y3, zishuo.wang.x3\}@elms.hokudai.ac.jp}
 {\tt\small ling@ist.hokudai.ac.jp}
}}

\begin{document}

\maketitle
\thispagestyle{empty}
\pagestyle{empty}

\begin{abstract}

Vision-language models have shown strong potential for social robot navigation by leveraging rich semantic understanding of complex environments and human behaviors. However, large scale VLMs are difficult to deploy on resource-constrained robotic platforms, while lightweight VLMs often lack sufficient social reasoning capability. To address this problem, we propose SOPD-SocialNav, a selective on-policy distillation (SOPD) method that transfers social navigation knowledge from a large teacher VLM to a lightweight student VLM. SOPD introduces an entropy-based token selection mechanism that uses teacher uncertainty to identify socially informative decision tokens, while suppressing gradients from low-entropy tokens corresponding to trivial navigation states. A temperature-controlled Jensen-Shannon divergence objective is then used to align the student and teacher distributions on the selected tokens. Experiments on the SNEI and MUSON benchmarks demonstrate that SOPD consistently outperforms supervised fine-tuning, off-policy distillation, and standard on-policy distillation baselines in action prediction, perception consistency, and reasoning consistency. Real-world deployment on a Scout Mini robot further shows that the distilled model can generate more socially appropriate navigation behaviors in conversational and queuing scenarios. These results suggest that SOPD is an effective strategy for building lightweight yet socially aware VLM-based navigation systems.

\end{abstract}

\section{INTRODUCTION}

When robots perform navigation tasks, many scenarios require human-centered decision-making. For example, in outdoor campuses, parks, libraries, and other environments, robots often encounter large numbers of pedestrians. In these situations, if a robot only considers obstacle avoidance or motion efficiency, it may cause a sense of pressure or discomfort for nearby pedestrians. Therefore, human-centered social navigation is essential for deploying robots in human-centered environments.

Traditional methods for social navigation mainly include the social force model~\cite{helbing1995social}, proxemics-based methods~\cite{bilen2024social}, velocity obstacles~\cite{fiorini1998motion}, DWA~\cite{fox1997dynamic} with social costs, imitation learning~\cite{yildirim2024learning}, and reinforcement learning~\cite{chen2017socially}. Although these policies can learn socially compliant behaviors, they typically rely on direct observation-to-action mappings, with the semantic and social reasoning required to interpret human intentions, group dynamics, and social norms only implicitly captured. This limitation has motivated growing interest in leveraging vision-language models (VLMs) for social navigation, as their high-level semantic understanding and reasoning capabilities offer a promising way to explicitly account for social context in navigation decisions.

Recent advances in VLMs have demonstrated that large scale VLMs possess strong semantic understanding and commonsense reasoning capabilities, enabling them to assess complex social costs and support navigation decisions that better align with human expectations~\cite{song2024vlm}. However, due to computational resource limitations, such models are difficult to deploy directly on edge devices.  A practical alternative is to use large VLMs as teachers and distill their socially informed decision-making knowledge into lightweight student VLMs for efficient real-time deployment. 
However, traditional on-policy distillation (OPD)\cite{song2026survey} usually adopts Reverse Kullback–Leibler (KL) divergence~\cite{gu2024minillm} as the optimization objective, while Reverse KL divergence has an evident mode-seeking~\cite{murphy2012machine} property. It encourages the student model to fit only the action with the highest probability in the teacher distribution, which reduces action diversity and easily leads to mode collapse. This problem becomes more severe when the teacher distribution exhibits high-entropy or multimodal characteristics, as the gradient signals during training may become more unstable. 
In addition, traditional OPD typically computes the distillation loss over all tokens. However, navigation trajectories often contain routine motion states, such as moving continuously forward through an open corridor, in which both the teacher and student are highly confident about the appropriate action. The corresponding low-entropy tokens primarily reflect deterministic geometric motion rather than complex social interactions. Treating these tokens equally in the distillation objective can cause gradient updates to be dominated by abundant routine-motion examples, thereby reducing the relative emphasis on socially critical behaviors, such as yielding to pedestrians and avoiding pedestrian flows.

To address the above problems, we propose an entropy selective on-policy distillation method, which aims to transfer the social reasoning capability of a large teacher VLM to a lightweight student VLM. At the same time, it selectively suppresses interference from irrelevant and redundant information in navigation decision-making, encouraging the student model to focus on socially relevant cues and socially compliant navigation behaviors.

The main contributions of this work are summarized as follows:
\begin{itemize}
    \item We propose an entropy selective on-policy distillation (SOPD) method for VLM-based social navigation. The proposed approach identifies socially informative decision tokens through teacher uncertainty entropy estimation and selectively transfers knowledge from a large scale teacher VLM to a lightweight student VLM.
    \item We provide a systematic study of distillation paradigms for social navigation VLMs, including off-policy distillation, standard OPD, and the proposed method. We also investigate the performance of different OPD loss functions for fine-tuning VLMs in social navigation tasks. Results on the SNEI and MUSON benchmarks demonstrate the effectiveness of entropy guided knowledge transfer.
    \item We demonstrate successful deployment of the distilled model on a Scout Mini robot in real-world social navigation scenarios, showing improved compliance with human-aware navigation conventions.
\end{itemize}

\section{Related work}

\subsection{VLMs in Social Robot Navigation}

Recent advances in VLMs, particularly in visual understanding, commonsense reasoning, and contextual interpretation, have fostered growing interest in VLM-based social navigation. VLM-SocialNav~\cite{song2024vlm} uses a closed-source large scale model as a social reasoning module. By semantically understanding the scene and evaluating the social appropriateness of candidate robot actions, it provides social costs to the downstream planner, thereby generating navigation behaviors that better align with human expectations. However, it relies on a large scale GPT model accessed through an online API, which introduces substantial inference latency and hinders direct deployment on resource-constrained robotic platforms. Moreover, without fine-tuning on the target social navigation dataset, the VLM may struggle to adapt to domain-specific social behaviors and action preferences. 
To improve domain adaptation, several recent studies have explored fine-tuning VLMs on task-specific social navigation datasets. 
Social-LLaVA~\cite{payandeh2025social} constructs the SNEI dataset specifically for social navigation and fine-tunes LLaVA-v1.5-7B using this dataset, enabling the model to perform social navigation reasoning in terms of perception, prediction, reasoning, action, and explanation. AutoSpatial~\cite{kong2025autospatial} further enhances the spatial perception and reasoning capabilities of VLMs regarding pedestrians’ relative positions, motion directions, and interaction relationships through structured spatial grounding, automatically annotated VQA data, and a small amount of human annotation, thereby improving action decision accuracy in social navigation. However, such approaches often require substantial training resources.

To address the latency issue, recent studies have explored social navigation with lightweight VLMs. MAction-SocialNav~\cite{wang2025maction} formulates social navigation as a multi-action ranking problem and develops a lightweight VLM based on NVILA-2B, enabling the efficient representation and ranking of multiple reasonable navigation choices beyond single-action prediction. SocialNav-MoE~\cite{kawabata2025socialnav} and E-SocialNav~\cite{xiao2026socialnav} improve action decision-making capabilities of light-weight VLM through reinforcement learning. Despite these advances, existing methods primarily improve lightweight VLMs through task-specific architectural designs or direct policy optimization, without fully exploiting the socially informed knowledge available in large scale VLMs to guide lightweight models.

\subsection{Knowledge Distillation for VLMs}
Knowledge distillation aims to transfer knowledge from a large scale teacher model to a more lightweight student model, enabling the student model to maintain performance close to that of the teacher model with lower inference cost. Early knowledge distillation mainly supervised the student model through soft labels~\cite{hinton2014distilling}, allowing it to learn the inter-class relationships contained in the output distribution of the teacher model. With the development of deep models, knowledge distillation has gradually expanded to intermediate feature alignment~\cite{hinton2014distilling}, attention transfer~\cite{zagoruyko2017paying}, representation space constraints, and generative sequence distillation~\cite{kim2016sequence}. In recent years, with the development of large language models (LLMs) and VLMs, knowledge distillation has further been used to compress the perception, reasoning, and instruction-following capabilities of large scale models, enabling smaller models to adapt more efficiently to complex downstream tasks~\cite{hsieh2023distilling}.

In VLMs, knowledge distillation is commonly used to transfer the cross-modal understanding and reasoning capabilities of a teacher model to a student model~\cite{jang2025vl2lite}. Common methods include SFT based on teacher-generated output sequences~\cite{wang2023self}, soft-label distillation based on logits or probability distributions~\cite{hinton2014distilling}, and feature-level distillation that incorporates intermediate vision-language representations. These methods can effectively improve the performance of small models in tasks such as visual question answering, image captioning, and embodied decision-making, while reducing computational overhead during deployment. However, most traditional distillation methods still belong to the off-policy paradigm, where the student model mainly learns from offline data or fixed outputs generated by the teacher model, response sequences generated by its current policy. This mismatch between the training distribution and the states encountered during deployment can cause errors to accumulate when the student deviates from the teacher-generated response sequences, limiting its ability to make reliable decisions in dynamically evolving social interactions.

To reduce the distribution mismatch between training and inference stage, OPD has gradually attracted increasing attention~\cite{gu2024minillm}. Unlike traditional off-policy distillation, OPD first allows the student model to generate output sequences according to its own policy, and then the teacher model provides token-level supervision~\cite{li2026rethinking} based on the context generated by the student. Therefore, the student model can learn the behavioral preferences of the teacher model under the state distribution that it may realistically encounter. This paradigm is particularly suitable for sequence generation and decision-making tasks, because during inference, the model needs to continuously rely on its previously generated results to make subsequent decisions. However, existing OPD methods usually apply a uniform distillation objective to all tokens and rarely distinguish the importance of different tokens to the final decision.

\section{Method}
\begin{figure}
    \centering
    \includegraphics[width=\linewidth]{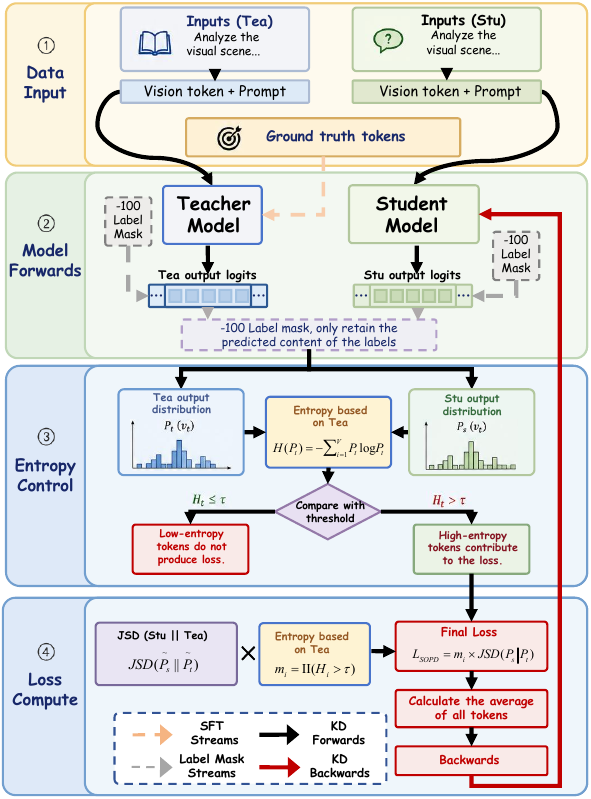}
    \caption{\textbf{Overview of SOPD.} The proposed SOPD framework takes text--image pairs from social navigation datasets as input. After supervised fine-tuning, the teacher VLM is frozen, while the student interacts with the environment and generates responses under its current policy. The frozen teacher then evaluates the student responses to obtain token-level probability distributions. Based on these distributions, Entropy-controlled token selection then identifies socially informative decision tokens for selective distillation. The resulting SOPD loss is backpropagated to update the student model.}
    \label{fig:method}
\end{figure}

\subsection{Overview}

The objective of this work is to distill the socially informed perception and decision-making knowledge of a large teacher VLM into a lightweight student model, enabling efficient deployment while maintaining strong social navigation performance. To achieve this goal, we propose \textbf{Selective On-Policy Distillation (SOPD)}, an on-policy knowledge distillation method that selectively transfers socially relevant knowledge from a frozen teacher VLM to a student VLM.

As illustrated in Fig.~\ref{fig:method}, the method consists of four stages.
First, the teacher VLM is supervised fine-tuned on the social navigation dataset and then frozen throughout distillation. Second, conditioned on the current visual observation and task prompt, the student model autoregressively samples a response sequence from its current policy, where each token is generated based on the multimodal input and all previously generated tokens. Third, the student-generated response sequences are evaluated by the frozen teacher VLM, which produces token-level probability distributions. Finally, a selective distillation mechanism identifies socially informative decision tokens and performs knowledge transfer only on the selected tokens. 
Unlike conventional off-policy distillation methods that rely on static datasets, SOPD supervises the student model under its own state distribution. Furthermore, instead of treating all tokens equally, the proposed SOPD method focuses supervision on socially ambiguous decision regions, enabling more effective transfer of social navigation knowledge.

\subsection{Selective On-Policy Distillation (SOPD)}

\noindent\textbf{On-Policy Distillation.}
Most existing VLM distillation methods rely on offline demonstrations collected from expert policies. However, the state distribution encountered by the student during deployment often differs from the distribution represented in the training dataset, leading to distribution mismatch and degraded performance.
To mitigate this issue, we adopt an OPD paradigm. Specifically, the teacher model is first fine-tuned through supervised instruction tuning (SFT) on social robot navigation datasets and then frozen during distillation. Then, responses $y$ are generated by the current student model using user prompt $x$.
\begin{equation}
y \sim \pi_S(x),
\end{equation}
where $\pi_S$ denotes the student model.

For each generated response, the frozen teacher model and the current student model are queried to obtain token-level output distributions. Let $P_t(y)$ and $P_s(y)$ denote the teacher and student probability distributions over the vocabulary at each response token, respectively.
To expose the fine-grained relative probabilities encoded in the teacher outputs, we apply temperature scaling to both distributions:

\begin{equation}
\tilde{P}_t=\mathrm{softmax}\left(\frac{z_t}{\lambda}\right),
\end{equation}

\begin{equation}
\tilde{P}_s=\mathrm{softmax}\left(\frac{z_s}{\lambda}\right),
\end{equation}

where $z_t$ and $z_s$ are the logits produced by the teacher and student models, respectively, and $\lambda$ denotes the temperature parameter.

Knowledge transfer is performed by minimizing the Jensen-Shannon Divergence (JSD) between the softened distributions:

\begin{equation}
\mathcal{L}_{JSD}(\tilde{P}_t \parallel \tilde{P}_s) = \frac{1}{2} D_{KL}(\tilde{P}_t \parallel M) + \frac{1}{2} D_{KL}(\tilde{P}_s \parallel M),
\end{equation}
where $M = \frac{1}{2}(\tilde{P}_t + \tilde{P}_s)$, $D_{KL}$ denotes the KL divergence.

Compared with conventional off-policy distillation, the proposed strategy allows the student model to receive supervision under states induced by its own behavior, thereby reducing distribution mismatch and improving policy robustness.

\noindent\textbf{Selective Token Distillation.}
In social navigation environments, not all generated tokens contribute equally to decision making. Many tokens correspond to straightforward navigation behaviors and provide limited additional supervision. In contrast, socially ambiguous situations, such as ``yielding to pedestrians'', ``selecting avoidance directions'', or ``negotiating passage through crowded spaces'', often contain richer social reasoning information.

To identify these informative decision regions, we estimate the uncertainty of the teacher prediction at each decoding step using entropy:
\begin{equation}
H_i=\sum_{k}\tilde{P}_t(i)\log \tilde{P}_t(i),
\end{equation}
where $\tilde{P}_t(i)$ denotes the teacher probability distribution assigned to token $i$.

Based on the entropy value, a binary selection mask is defined as
\begin{equation}
m_i=\mathbb{I}\left(H_i > \tau\right),
\end{equation}
where $\tau$ denotes the entropy threshold and $\mathbb{I}(\cdot)$ is the indicator function.
Ensuring only the selected tokens participate in distillation:

\begin{equation}
\mathcal{L}_{SOPD}=\sum_i m_i \cdot \mathcal{L}_{JSD}(\tilde{P}^{i}_t \parallel \tilde{P}^{i}_s).
\end{equation}

This selective mechanism suppresses gradient contributions from trivial predictions and encourages the student model to focus on socially informative navigation decisions. The resulting selective distillation loss serves as the final training objective of SOPD. During distillation, all token-level supervisory signals are provided by the SFT-trained and frozen teacher model.

\section{Experiment}

\subsection{Experiment Settings}

\noindent\textbf{Dataset Benchmark.}
To comprehensively evaluate the performance of our SOPD method, we compare with several competitive knowledge distillation baselines, including off-policy knowledge distillation methods, namely SKD~\cite{xu2025speculative} and VL-KD~\cite{jang2025vl2lite}, and on-policy knowledge distillation methods, namely OPD~\cite{agarwal2024policy} and OPSD~\cite{zhao2026self}. All experiments were conducted on the SNEI~\cite{payandeh2025social} and MUSON~\cite{liu2025muson} datasets, respectively. For each dataset, the data were randomly split into training and test sets with a ratio of 8:2. For a fair comparison, all methods used Qwen3-VL-8B~\cite{bai2025qwen3} as the large scale teacher model and Qwen2.5-VL-3B as the lightweight student VLM. The student model was trained for 5 epochs with a learning rate of $1\times10^{-5}$, and a cosine learning rate scheduler with a warmup ratio of 0.1 was adopted. The temperature parameter $\lambda$ was set to $2.0$, and the entropy threshold $\tau$ was set to $1.5$. All remaining hyperparameters were kept consistent across methods for a fair comparison. The weighted Jensen--Shannon divergence used in our objective follows the formulation of OPD~\cite{agarwal2024policy}. All experiments were conducted with eight NVIDIA A100 GPUs.

\noindent\textbf{Validation Metrics.}
To comprehensively evaluate the quality of the social navigation outputs generated by the model, we employ a diverse set of metrics: We report Action Accuracy (Acc). Specifically, given an output action $y$ and a ground truth action $g$, Action Accuracy is defined as
\begin{equation}
\text{Acc}=\frac{1}{N}\sum_{i=1}^{N}\mathbf{1}\!\left(y_i = g_i\right),
\end{equation}
where $\mathbf{1}(\cdot)$ denotes the indicator function, which equals $1$ if the predicted action exactly matches the ground truth action and $0$ otherwise.
To explicitly evaluate semantic alignment in social environments and reasoning processes, we employ Perception Cosine Similarity (Per-cos) and Reasoning Cosine Similarity (Rea-Cos). These two metrics use a pre-trained Sentence BERT model to compute the cosine similarity between the sentence embeddings of the generated text and the reference text, providing a robust measure of semantic similarity in the embedding space.

\subsection{Main Results}

\begin{table}[t]
\centering
\caption{Comparison with state-of-the-art off-policy and on-policy knowledge distillation methods on the SNEI and MUSON datasets. The best results are highlighted in bold.}
\label{tab:main_results}
\resizebox{0.5\textwidth}{!}{
\begin{tabular}{lcccccc}
\toprule
\multirow{2}{*}{Method} 
& \multicolumn{3}{c}{SNEI} 
& \multicolumn{3}{c}{MUSON} \\
\cmidrule(lr){2-4} \cmidrule(lr){5-7}
& Acc$\uparrow$ & Per-cos$\uparrow$ & Rea-cos$\uparrow$ 
& Acc$\uparrow$ & Per-cos$\uparrow$ & Rea-cos$\uparrow$ \\
\midrule
SFT 
& 0.692 & 0.834 & 0.767 
& 0.811 & 0.810 & 0.803 \\

\midrule
\multicolumn{7}{c}{\textit{Off-policy Knowledge Distillation}} \\
\midrule
SKD~\cite{xu2025speculative} 
& 0.831 & 0.828 & 0.819 
& 0.932 & 0.936 & 0.934 \\

VL-KD~\cite{jang2025vl2lite}
& 0.799 & 0.777 & 0.784 
& 0.916 & 0.910 & 0.897 \\

\midrule
\multicolumn{7}{c}{\textit{On-policy Knowledge Distillation}} \\
\midrule
OPD~\cite{agarwal2024policy}
& 0.792 & 0.795 & 0.797 
& 0.886 & 0.920 & 0.896 \\

OPSD~\cite{zhao2026self}
& 0.863 & 0.851 & 0.872 
& 0.895 & 0.907 & 0.903 \\

SOPD (Ours) 
& \textbf{0.889} & \textbf{0.901} & \textbf{0.894} 
& \textbf{0.953} & \textbf{0.938} & \textbf{0.955} \\

\bottomrule
\end{tabular}
}
\end{table}

Table~\ref{tab:main_results} presents the main comparison results of SOPD against state-of-the-art off-policy and on-policy knowledge distillation methods on the SNEI and MUSON datasets. Overall, our SOPD method consistently achieves the best performance across all evaluation metrics on both datasets. On the SNEI dataset, SOPD obtains an Action-Acc of 0.889, a Per-cos of 0.901, and a Rea-cos of 0.894, outperforming the strongest baseline OPSD by 0.026, 0.050, and 0.022, respectively. Compared with the SFT baseline, SOPD brings more substantial improvements, especially in Action-Acc, increasing the score from 0.692 to 0.889. This indicates that our method can effectively transfer the social reasoning capability of the teacher model to the lightweight student model and improve both action prediction and reasoning quality.

Similar trends can also be observed on the MUSON dataset. SOPD achieves the highest Action-Acc of 0.953, Per-cos of 0.938, and Rea-cos of 0.955. Compared with the best-performing baseline on each metric, SOPD improves Action-Acc over SKD by 0.021, Per-cos over SKD by 0.002, and Rea-cos over SKD by 0.021.

Among all compared methods, off-policy distillation methods such as SKD and VL-KD improve over SFT in most cases, showing that transferring knowledge from a stronger teacher model is beneficial for social navigation. However, their performance is still lower than SOPD, suggesting that relying only on fixed teacher-generated data or offline supervision may be insufficient to address the distribution mismatch between training and inference. On-policy methods such as OPD and OPSD further consider student-generated response sequences, but they still apply distillation objectives uniformly to tokens without explicitly distinguishing high-uncertainty decision-critical tokens. In contrast, SOPD introduces entropy-controlled token filtering and JSD-based distillation, allowing the student model to focus more on informative tokens that are closely related to social decision-making. This leads to more stable and effective knowledge transfer, resulting in state-of-the-art performance.

\begin{figure}[t!]
    \centering
    \includegraphics[width=\linewidth]{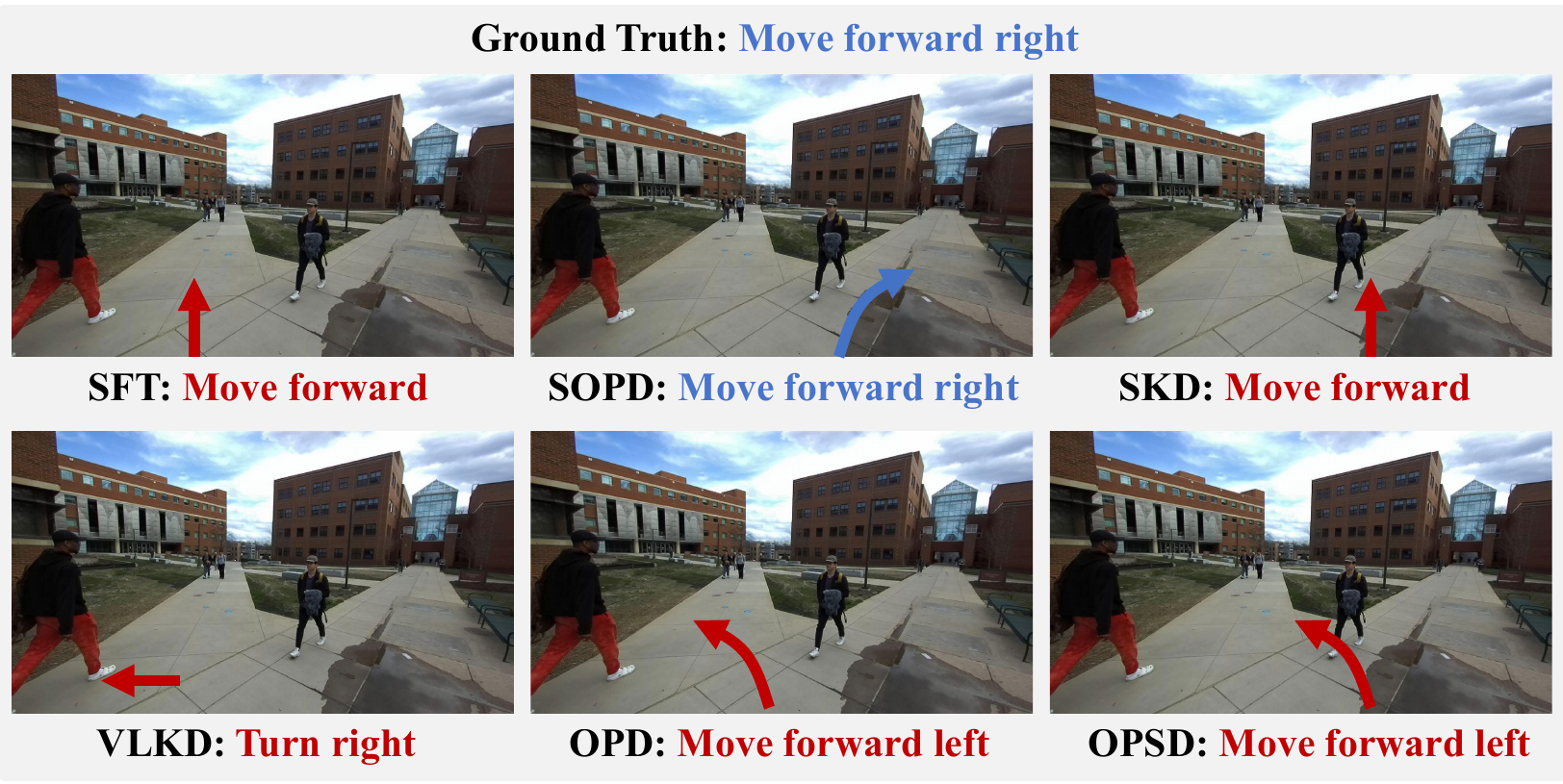}
    \caption{Visual comparison with other methods. SOPD can better capture the spatial relationship between the robot and pedestrians and generate more accurate socially compliant navigation decisions.}
    \label{fig:visual}
\end{figure}

As shown in Fig.~\ref{fig:visual}, SOPD correctly predicts the ground truth action, \textit{move forward right}, while the other baselines produce incorrect or socially less appropriate actions. This result indicates that SOPD can better capture the spatial relationship between the robot and pedestrians and generate more accurate socially compliant navigation decisions.

\subsection{Ablation Studies}

\begin{table}[t]
\centering
\caption{Main ablation studies of the proposed SOPD method.}
\label{tab:main_ablation}
\resizebox{\linewidth}{!}{
\begin{tabular}{lccccc}
\toprule
Method & $m_i$ & $\mathcal{L}_{JSD}$ & Acc & Per-cos & Rea-cos \\
\midrule
SFT 
& $\times$ & $\times$ 
& 0.811 & 0.810 & 0.803 \\

w/o $m_i$ 
& $\times$ & $\checkmark$ 
& 0.893 & 0.892 & 0.884 \\

w/o $\mathcal{L}_{JSD}$ 
& $\checkmark$ & $\times$ 
& 0.836 & 0.834 & 0.829 \\

Ours 
& $\checkmark$ & $\checkmark$ 
& \textbf{0.953} & \textbf{0.938} & \textbf{0.955} \\
\bottomrule
\end{tabular}
}
\end{table}

To examine the contribution of each component in SOPD, we evaluate the effects of the Jensen--Shannon divergence (JSD) objective and the entropy-based token filter, as summarized in Table~\ref{tab:main_ablation}. Compared with the baseline model, using JSD alone improves Action-Acc from 0.811 to 0.893, Per-cos from 0.810 to 0.892, and Rea-cos from 0.803 to 0.884. This result indicates that distribution-level alignment provides more effective supervision than the original distillation objective, improving both action prediction and reasoning consistency.

Using only the entropy filter also consistently improves performance, yielding 0.836 Action-Acc, 0.834 Per-cos, and 0.829 Rea-cos. Although these gains are smaller than those obtained with JSD alone, they suggest that prioritizing high-entropy tokens helps reduce the influence of routine low-entropy motion tokens and places greater emphasis on socially informative decision states.

Combining JSD with the entropy filter achieves the best overall performance, reaching 0.953 Action-Acc, 0.938 Per-cos, and 0.955 Rea-cos. Relative to JSD alone, the full SOPD model further improves Action-Acc, Per-cos, and Rea-cos by 0.060, 0.046, and 0.071, respectively. These results demonstrate that the two components are complementary: JSD aligns the teacher and student output distributions, while entropy-based selection concentrates supervision on socially informative tokens. Together, they facilitate more effective transfer of socially informed decision-making knowledge from the teacher model to the student model.

\subsection{Ablation Studies on Divergence Functions}

\begin{table}[t]
\centering
\caption{Comparison results with different divergence functions.}
\label{tab:divergence_ablation}
\resizebox{\columnwidth}{!}{
\begin{tabular}{lccc}
\toprule
Divergence & Acc$\uparrow$ & Per-cos$\uparrow$ & Rea-cos$\uparrow$ \\
\midrule
Reverse KL 
& 0.885 & 0.837 & 0.824 \\

Forward KL 
& 0.853 & 0.856 & 0.835 \\

Weighted JSD 
& 0.861 & 0.826 & 0.843 \\

Temperature JSD 
& \textbf{0.953} & \textbf{0.938} & \textbf{0.955} \\
\bottomrule
\end{tabular}
}
\end{table}

\begin{figure}[t]
\centering
\includegraphics[width=0.5\textwidth]{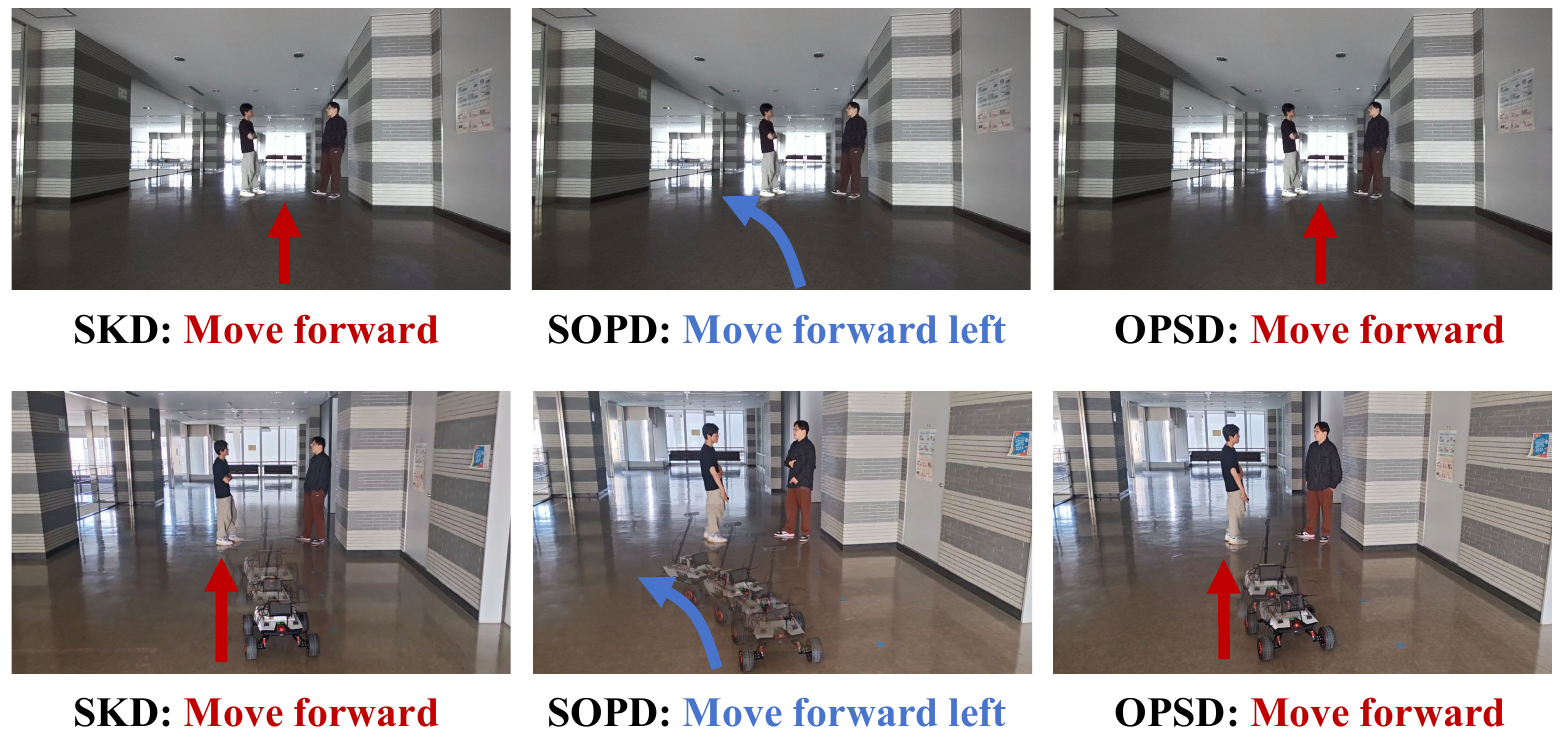}
\caption{Qualitative comparison in a real-world conversational scenario. Two pedestrians are engaged in a face-to-face conversation.}
\label{fig:realcase1}
\vspace{0.02em}
\centering
\includegraphics[width=0.5\textwidth]{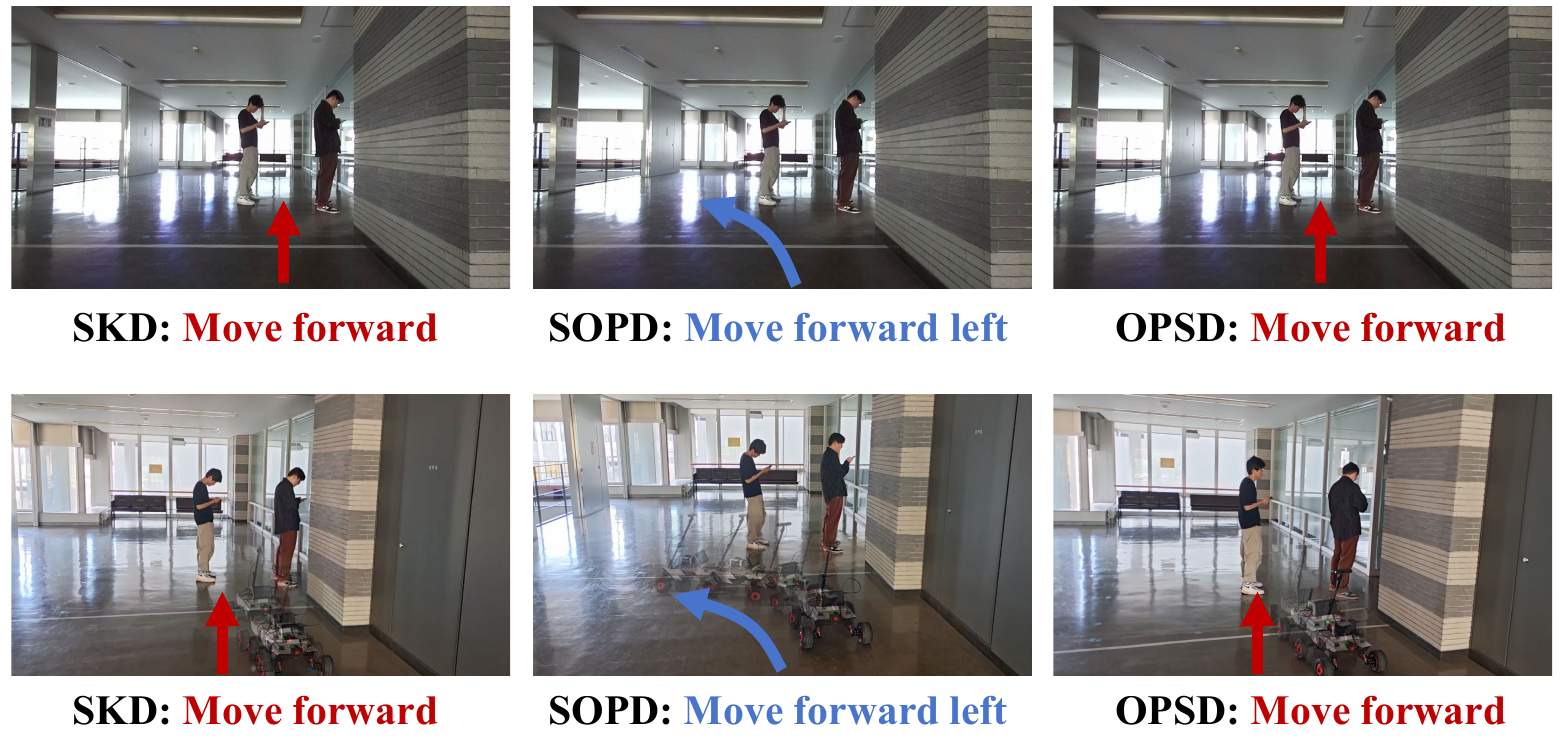}
\caption{Qualitative comparison in a real-world queuing scenario. Two pedestrians are standing in line and waiting in front of the robot.}
\label{fig:realcase2}
\end{figure}
To investigate the effect of different divergence functions in the distillation objective, we conduct ablation studies by replacing the proposed temperature JSD with reverse KL, forward KL, and weighted JSD. The results are shown in Table~\ref{tab:divergence_ablation}. Overall, temperature JSD achieves the best performance across all evaluation metrics, obtaining an Action-Acc of 0.953, a Per-cos of 0.938, and a Rea-cos of 0.955.

Compared with reverse KL, temperature JSD improves Action-Acc from 0.885 to 0.953, Per-cos from 0.837 to 0.938, and Rea-cos from 0.824 to 0.955. This suggests that Reverse KL, although effective for fitting high-probability teacher outputs, may suffer from its mode-seeking property and thus fail to preserve sufficient action diversity in social navigation scenarios. Forward KL achieves slightly better Per-cos and Rea-cos than Reverse KL, but its Action-Acc decreases to 0.853, indicating that directly encouraging mode-covering behavior may introduce uncertainty into action prediction.

Weighted JSD achieves a more balanced result than KL-based objectives, especially improving Rea-cos to 0.843. However, its overall performance is still clearly lower than temperature JSD. This indicates that simply using JS-based distribution matching is insufficient, and temperature control plays an important role in stabilizing the distillation process and improving knowledge transfer. By introducing temperature-controlled JSD, our method can better align the student distribution with the teacher distribution while maintaining stable gradients during optimization. These results demonstrate the effectiveness of the proposed temperature JSD objective for transferring social reasoning capability in SOPD.

\subsection{Real-world Experiments}

We further deployed the SOPD-distilled model on a Scout Mini robot to evaluate its practical feasibility for real-world social navigation. The model achieves an inference throughput of 0.984 FPS. When accounting for the complete end-to-end pipeline, including network communication and image upload/download, the average latency is 2.74 seconds per sample, corresponding to an effective system throughput of 0.365 FPS.

As shown in Fig.~\ref{fig:realcase1}, two pedestrians are engaged in a face-to-face conversation, forming an interaction space that the robot should avoid entering. In this scenario, directly moving forward may cause the robot to approach the pedestrians too closely, interrupt their conversation, and create a sense of pressure. Given the same egocentric observation, SKD and OPSD both output \textit{move forward}, which is geometrically feasible but socially less appropriate. In contrast, SOPD outputs \textit{move forward left}, allowing the robot to bypass the conversational group from the side and maintain a larger social distance. This result indicates that SOPD can better recognize socially sensitive situations and generate navigation actions that are more consistent with human-aware navigation norms.

As shown in Fig.~\ref{fig:realcase2}, we further evaluate the model in a queuing scenario, where two pedestrians are standing in line or waiting in the robot's forward direction. In such a situation, a socially compliant robot should avoid directly approaching the queue from behind, since moving straight forward may reduce the distance to the pedestrians and create a sense of pressure. Instead, the robot is expected to adjust its trajectory and pass from the side while preserving the queue space. Given the same egocentric observation, SKD and OPSD both predict \textit{move forward}, which may cause the robot to approach the pedestrians too closely and violate human-aware navigation conventions. In contrast, SOPD predicts \textit{move forward left}, enabling the robot to bypass the waiting pedestrians from the side. This result shows that SOPD can better recognize socially sensitive spatial arrangements, such as queues, and generate navigation actions that maintain a larger social distance and better conform to social norms.

\section{CONCLUSIONS}

In this paper, we proposed SOPD-SocialNav, an entropy-selective extension of on-policy distillation for vision-language social robot navigation. Based on the OPD training paradigm, SOPD transfers socially informed perception and decision-making knowledge from a large teacher VLM to a lightweight student VLM, supporting efficient deployment on resource-constrained robotic platforms.
SOPD supervises the student using responses generated under its current policy and introduces teacher-entropy-based token selection to prioritize socially informative and decision-critical tokens. By reducing gradient contributions from low-entropy routine-motion tokens, the proposed method concentrates distillation on the social interactions most relevant to socially compliant navigation.

Experimental results on the SNEI and MUSON benchmarks demonstrate that SOPD consistently outperforms supervised fine-tuning, off-policy distillation, and standard OPD baselines in action prediction, perception consistency, and reasoning consistency. Ablation studies further show that both entropy-based token selection and temperature-controlled JSD contribute to the final performance, and their combination enables more stable and effective knowledge transfer. These results indicate that SOPD is a promising strategy for building lightweight yet socially aware VLM-based navigation systems.

In future work, we plan to further investigate adaptive entropy threshold selection, extend SOPD to longer-horizon navigation trajectories, and evaluate the method in more diverse real-world human-robot interaction scenarios.

\bibliographystyle{IEEEtran}
\bibliography{ref}

\end{document}